\documentclass{article}
\usepackage{spconf,amsmath,graphicx}
\usepackage{ragged2e} 
\usepackage{booktabs,makecell, multirow, tabularx}
\usepackage[breaklinks]{hyperref}

\title{Dynamic Frequency Domain Graph Convolutional Network for Traffic  Forecasting}
\name{Yujie Li$^{1,2}$, Zezhi Shao$^{1,2}$, Yongjun Xu$^{1,2}$, Qiang Qiu$^{1,2}$, Zhaogang Cao$^{3}$, Fei Wang$^{1,2}$\sthanks{corresponding author: wangfei@ict.ac.cn\\  
		This work is  supported by NSFC No. 62372430 and the Youth Innovation Promotion Association CAS No.2023112.}}
\address{$^{1}$Institute of Computing Technology, Chinese Academy of Sciences, Beijing, China\\$^{2}$University of Chinese Academy of Sciences, Beijing, China\\$^{3}$Academy of Cyber, Beijing, China}

\begin{document}
%
\maketitle
\begin{abstract}
Complex spatial dependencies in transportation networks make traffic prediction extremely challenging. Much existing work is devoted to learning dynamic graph structures among sensors, and the strategy of mining spatial dependencies from traffic data, known as data-driven, tends to be an intuitive and effective approach. However, Time-Shift of traffic patterns and noise induced by random factors hinder data-driven spatial dependence modeling. In this paper, we propose a novel dynamic frequency domain graph convolution network (DFDGCN) to capture  spatial dependencies. Specifically, we mitigate the effects of time-shift by Fourier transform, and introduce the identity embedding of sensors and time embedding when capturing data for graph learning since traffic data with noise is not entirely reliable. The graph is combined with static predefined and self-adaptive graphs during graph convolution to predict future traffic data through classical causal convolutions. Extensive experiments on four real-world datasets demonstrate that our model is effective and outperforms the baselines.
\end{abstract}
\section{Introduction}
\label{sec:intro}

With the increase of global urbanization and population concentration, complex transportation environments have become a major challenge for urban management, leading to the urgent need for Intelligent Transportation Systems (ITS)\cite{DBLP:journals/soco/JabbarpourZKSC18} about AI\cite{xu2021artificial} to enhance the population carrying capacity of cities. 

Traffic prediction is an indispensable part of ITS, which aims to accurately predict future traffic data by learning historical traffic states and patterns\cite{DBLP:journals/tvcg/LeeKJKMEK20} obtained from sensors.
 However, due to the high correlation between sensors, future information of each sensor relies not only on its own historical information but also on the historical data of other sensors. Spatio-temporal graph neural networks are proposed to capture not only the temporal dependence of traffic data inherent in each sensor, but also to mine the spatial dependence among sensors, and this class of methods has been proven to outperform general time series forecasting\cite{yu2023dsformer} methods such as ARIMA\cite{williams2003modeling} and LSTM\cite{ma2015long} with the main enhancement coming from mining the reliable spatial dependencies among sensors.

Mining effective spatial dependencies, also known as graph structures, has been an important challenge in traffic prediction. Early schemes such as DCRNN\cite{li2018} relies on prior spatial information to calculate predefined graph structures, GWNet\cite{wu2019graph} self-adaptively learns graph structures with learnable parameters. STGCN\cite{yu2018spatio} , AGCRN\cite{bai2020adaptive}, and MTGNN\cite{wu2020connecting} all apply or improve upon these static graphs to achieve outstanding enhancements.

 In subsequent research, researchers find that the spatial dependence of different times is not the same due to the complex urban transportation environment. ASTGNN\cite{guo2021learning} and DMSTGCN\cite{han2021dynamic} apply learnable parameters at each timestamp such as day of week or hour of day to self-adaptively capture spatial dependencies over different time periods. 
 In contrast, the approach of applying changing historical traffic data to model dynamic graphs is called data-driven. Since traffic patterns can always change randomly, data-driven approaches that can adjust the graph structure based on historical information during the testing phase \cite{shin2022pgcn} tend to be more effective than self-adaptive approaches that rely purely on learnable parameters. STFGCN\cite{li2021spatial}, DGCRN\cite{li2023dynamic}, and DSTAGNN\cite{lan2022dstagnn} all utilize self-attention scores of time domain information or the spatial proximity between traffic data in time domain to construct the graph structures.

However, we argue that data-driven modeling of spatial dependence faces two problems: Time-Shift and Data Noise. 

\begin{figure}[htb] 	 		
	\centering 		
	\centerline{\includegraphics[width=7.5cm]{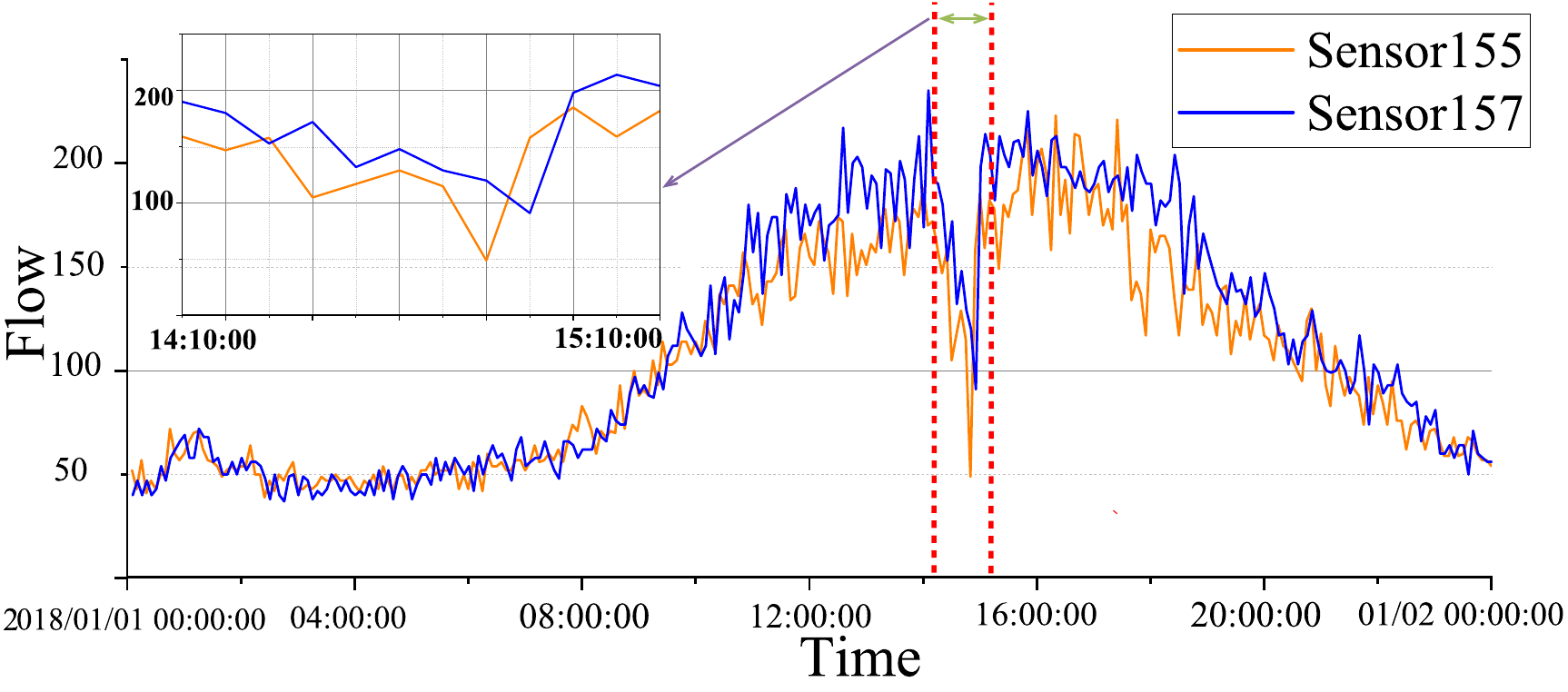}} 	
	\vspace{-3mm} 		
	\caption{Time-shift in Traffic Forecasting} 		
	\label{timeshift} 		
	\vspace{-3mm} 
\end{figure}

Time-shift is a common problem in urban transportation. For instance,  during urban commuting,  when people come off duty from industrial areas, the congested traffic can take half an hour or more to reach residential areas. Time-shift of traffic patterns due to similar situations is common in modern cities because of urban functional zones, and it can result in tidal wave-like delay variations between interrelated traffic data as shown in Fig. \ref{timeshift} .
\vspace{-3mm} 

\begin{figure}[htb]
	\vspace{-3mm}
	\centering
	\centerline{\includegraphics[width=4.5cm]{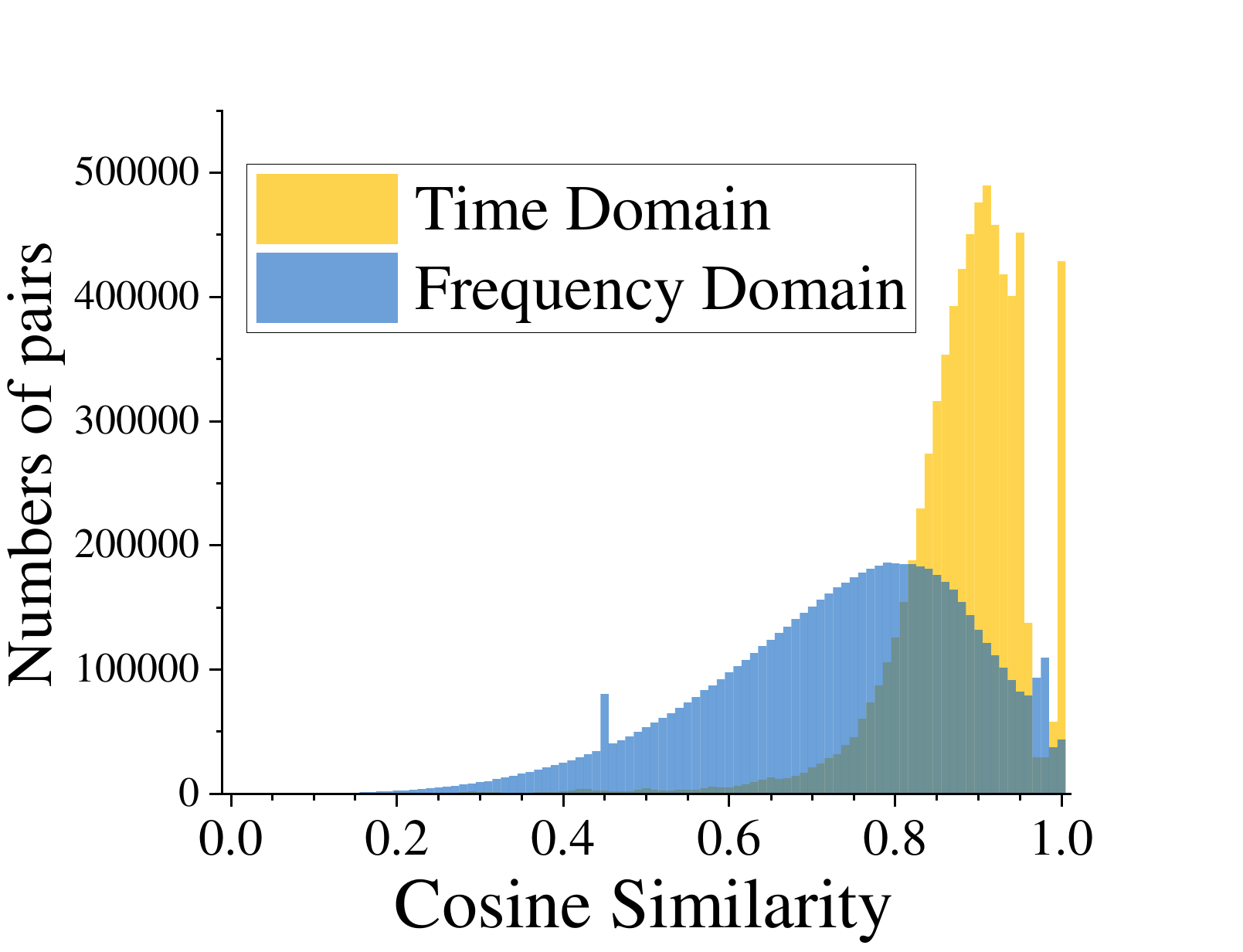}}
	\vspace{-3mm}
	\caption{Cosine similarity of traffic flow in time and frequency domains within one weak}
	\label{cos}
	\vspace{-3mm}
\end{figure}

However, time-shift can invalidate spatial proximity measures, which are widely applied to capture spatial dependence of traffic data. It's because schemes such as Euclidean distance and cosine similarity are computed in aligned dimensions, but changes in interrelated traffic data affected by time shifts do not occur at the same timestamp, which raises difficulties in examining proximity to capture the correlation between roads in time domain. Fig.\ref{cos} represents the cosine similarity between traffic flows at the same observation time in one week, and we can observe  that the similarity between traffic flows in time domain exhibits low differentiation and stacking. This is due to the misalignment of the temporal dimension caused by time-shift, which makes the traffic flows on related roads have similar similarity to those on uncorrelated roads, which makes it difficult to explore valid spatial dependencies through proximity.

Noise in traffic data is an unavoidable problem as in general data-driven strategies, and is often triggered by emergencies such as traffic accidents or road construction. Therefore, traffic data with noise is not always reliable, and it is a realistic challenge to introduce additional information to reduce the effect of noise while learning the spatial relationships between sensors.

To solve the above problems, we propose a dynamic frequency domain graph convolutional network(DFDGCN). To mitigate the effect of time-shift, we propose to examine the traffic patterns through the frequency domain information of traffic data. As shown in Fig.\ref{cos}, the frequency domain information between sensors exhibits better differentiation, it is because even if the traffic information is time-shifted, relying on the time-shift property of the Fourier transform, the frequency domain components of traffic data will still be in the same phase dimension, which will facilitate us to mine dynamic spatial dependence subsequently. In addition, we introduce identity embedding of sensors and time embedding when capturing traffic data for graph learning to minimize the effect of data noise. Specifically, we summarize the main contributions of our work as follows:
\vspace{-1mm}
\begin{itemize}
 \item We focus on the time-shift problem when mining spatial dependence of traffic data, and to the best of our knowledge we are the first to  mitigate it by frequency domain analysis with the property of Fourier transform.
 
\vspace{-2mm} 
\item We propose DFDGCN to learn dynamic graph structures among sensors  by combining traffic data with identity embedding and time embedding to cope with data noise.
\vspace{-2mm} 

\item Extensive experiments on four publicly available datasets demonstrate that our proposed model outperforms baselines, and ablation experiments witness the validity of our model.
\end{itemize}
\vspace{-5mm}
\section{METHODOLOGY}
\label{sec:format}
\vspace{-2mm}
\subsection{Fourier transformation}
\label{ft}
As described in the previous section, Time-Shift problem is a challenge for data-driven spatial dependence modeling, and we propose to address the problem by Fourier Transforming, which transfers traffic data to the frequency domain. 

Assuming the existence of traffic data $f(t)$ captured by sensors at a particular intersection, traffic data $f(t-t_{0})$ can be captured when the traffic flow  is delayed by time $t_{0}$ to reach the next intersection. In the process of spatial dependence modeling, learning the relationship between $f(t)$ and $f(t-t_{0})$ under the same time window is difficult or has a high computational overhead due to Time-Shift.

However, their corresponding Fourier transforms $F(\omega)$, $F_{t_{0}}(\omega)$ are related under the same dimension, according to the definition of Fourier transform:
\begin{flalign}
	F_(\omega) = \int_{-\infty}^{\infty}f(t)e^{-j\omega t}dt
\end{flalign}
\vspace{-4mm}
\begin{flalign}
	F_{t_{0}}(\omega) = \int_{-\infty}^{\infty}f(t-t_{0})e^{-j\omega t}dt
\end{flalign}
According to $t$ - $t_{0}$ = $c$, $F_{t_{0}}(\omega)$ can be continued to be expressed as:
\begin{flalign}
F_{t_{0}}&(\omega) = \int_{-\infty}^{\infty}f(c)e^{-j\omega (t_{0}+c)}dc \\ \notag
		&= [\int_{-\infty}^{\infty}f(c)\cdot e^{-j\omega c}dc]e^{-j\omega t_{0}}= F_(\omega)\cdot e^{-j\omega t_{0}}
\end{flalign}

Thus we prove that the traffic data in the frequency domain is represented in the same phase dimension, which solves the trouble caused by time-shift and means that learning the spatial dependencies between sensors will be more accurate and convenient.
\subsection{DFDGCN}
\label{dfdgcn}
\begin{figure}[htb]
	\vspace{-3mm}
	\centering
	\centerline{\includegraphics[width=8.5cm]{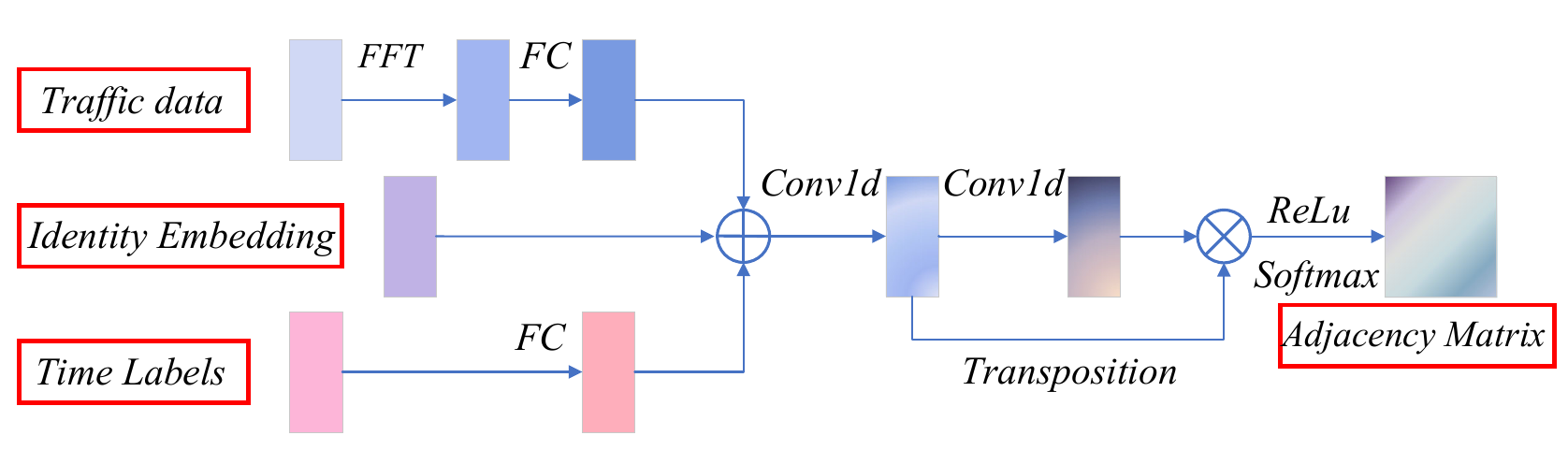}}
	\vspace{-3mm}
	\caption{The structure of proposed Frequency Domain Graph Module}
	\label{fdg}
	\vspace{-3mm}
\end{figure}

Urban traffic environments are complex and variable, and to alleviate the time-shift as well as noise problems in the data-driven dynamic graph modeling, we propose a frequency domain graph module described by Fig.\ref{fdg}. The core of DFDGCN is to update the dynamic adjacency matrix $A_{D}$ based on the traffic data observed at the current observation window.

According to section \ref{ft}, traffic data in the frequency domain receives less effect of time shift than that in the time domain. As a result, we transfer the traffic data $X_{t}$ at each observation window to the frequency domain by Fourier transform. 
\begin{flalign}
	F_{t} = FFT(X_{t})
\end{flalign}

In order to deal with the noise in traffic data, we introduce identity embedding\cite{DBLP:conf/cikm/ShaoZ00X22} and time embedding\cite{DBLP:journals/pvldb/ShaoZWWXCJ22} to increase additional information about the transportation network, so that we can easily extract the effective traffic patterns of each sensor and explore the spatial dependency among them:
\vspace{-1mm}
\begin{flalign} 	
	DE_{t} = W_{F,t}\cdot F_{t}|| E_{t}|| W_{T,t}\cdot (T^{W}_{t}||T^{D}_{t})
\end{flalign}
where $E_{t}$ represents learnable identity embedding of each sensor, $T^{W}_{t}$ $T^{D}_{t}$ denote the labels indicating day of week and hour of day, all $W$ are the learnable parameters for embedding the individual labels.

In addition, we apply a one-dimensional convolution layer with 1$\times$1 convolution kernel for further embedding, in order to learn the connections between the dimensions of $DE_{t}$. 

Finally, we apply a fully connected layer to learn $DE_{t}$ again to obtain directionality and then matrix multiply it with the transposition of $DE_{t}$. After the activation function and $Softmax$ we obtain the final adjacency matrix:
\vspace{-1mm}
\begin{flalign} 	 	
	A_{D}^{t} = Softmax(ReLU(DE_{t} W_{adj} DE^{T}_{t})) 
\end{flalign}

Consistent with previous work, we consider the dynamic frequency domain graph as a transition matrix for the hidden diffusion process. By combining it with the predefined graphs $P$ in DCRNN and the self-adaptive graph $A_{adt}$ in GWNet, we propose the following graph convolutional layer:  
\vspace{-1mm}
\begin{flalign} 	 	 	
	Z_{t} = \sum_{k=0}^{K}(P^{k}X_{t}W_{k,1}+A_{adt}^{k}X_{t}W_{k,2}+A_{D}^{t}X_{t}W_{k,3})  
\end{flalign}
For the processing of temporal information we adopt the same causal convolutional with residual network as in GWNet\cite{wu2019graph}.
\section{Experiments}
\subsection{Datasets}
We verify DFDGCN  on four commonly used large real-world datasets with tens of thousands of time steps and hundreds of sensors. The statistics are summarized in Table \ref{tab1}. PEMS-BAY is a traffic speed dataset and the others are traffic flow datasets, and the time steps for each dataset are obtained by sampling at five-minute intervals.

\begin{table}[htb] 	 
	    \vspace{-6mm}	 		 		
		\centering 		 		 		
		\caption{Statistics of Datasets}
		\label{tab1} 		 	 			 		
		\begin{tabular}{c|c|c|c|c} 			 				 		
			\toprule[1pt]
			\textbf{Type} & \textbf{Datasets} &\textbf{Time Steps} & \textbf{Nodes}& \textbf{Edges}  \\
			\toprule[1pt] 
			\textbf{Speed} & \textbf{PEMS-BAY} &52116 & 325& 2369  \\
			\hline \textbf{Flow} &  \textbf{PEMS03} &26208 & 358& 547  \\
			\hline \textbf{Flow} &  \textbf{PEMS07} &28224 & 883& 866  \\
			\hline \textbf{Flow} &  \textbf{PEMS08} &17856 & 170& 295  \\
			\toprule[1pt]
		\end{tabular} 	 			 	
	  \vspace{-10mm}	
\end{table}
\subsection{Baselines and Metrics}
We selected publicly available and classical baselines for traffic predicting, including recent traditional method HI\cite{cui2021historical}, representative deep learning methods GWNet\cite{wu2019graph}, DCRNN\cite{li2018}, AGCRN\cite{bai2020adaptive}, STGCN\cite{yu2018spatio}, MTGNN\cite{wu2020connecting}, DGCRN\cite{li2023dynamic}.  Due
to space limitations, we do not introduce each method in detail.We examine all baselines by three metrics, including mean absolute error (MAE), root mean square error (RMSE), and mean absolute percentage error (MAPE).

\vspace{-4mm}
\subsection{Experimental Setups}
The four datasets are divided into training, validation, and test sets in the ratio of 7:1:2. We predict future traffic data of 12 time steps with historical traffic data of length 12, and compare the performances of the 3rd, 6th, 12th and the average of the 12 timestamps, which are labeled @3, @6, @12 and Avg.. In our model, The dimensions of the Fourier transformed time series embedding as well as the identity embedding are both 10, and $T^{W}_{t}$$T^{D}_{t}$ in the temporal embedding are both 12 dimensions. The embedding size after 1-dimensional convolution is 30. Baselines as well as training strategies can be found in the publicly available benchmark\footnote{https://github.com/zezhishao/BasicTS}. The best experimental results will be marked boldly while the second will be underlined. The source code of DFDGCN  is available
\footnote{https://github.com/blisky-li/DFDGCN}.

\vspace{-3mm}
\subsection{Experimental Results}
\vspace{-2mm}
\begin{table*}
	\caption{Results of DFDGCN and Baselines for traffic forecasting on four real datasets}
	\label{res}
	\setlength{\tabcolsep}{1.8pt}
	\footnotesize
	\centering
		\begin{tabular*}{\linewidth}{@{}cc|cccc|cccc|cccc|cccc@{}}
			\hline
			\hline
			\multicolumn{2}{c|}{\textbf{Datasets}}&\multicolumn{4}{c|}{ \textbf{PEMS-BAY}}&\multicolumn{4}{c|}{ \textbf{PEMS03}}&   \multicolumn{4}{c|}{\textbf{PEMS07}}&\multicolumn{4}{c}{\textbf{ PEMS08}}\\
			\hline\hline
			 \textbf{Method} &  Metric& @3&@6&@12&Avg.& @3&@6&@12&Avg.& @3&@6&@12&Avg.& @3&@6&@12&Avg.\\
			\hline
			\multirow{3}{*}{ \textbf{HI}} & MAE&3.06&			3.06&			3.05&			3.05&			32.46&			32.45&			32.44&			32.45&			49.02&			49.04&			49.06&			49.03&			36.65&			36.66&			36.68&			36.66
			
			\\ 
			 & RMSE&7.05&			7.04&			7.03&			7.05&			49.78&			49.76&			49.75&			49.76&			71.15&			71.18&			71.21&			71.18&			50.44&			50.45&			50.46&			50.45
			  \\
			 &MAPE&6.85\%&			6.84\%&			6.83\%&			6.84\%&			30.58\%&			30.59\%&			30.63\%&			30.60\%&			22.73\%&			22.75\%&			22.79\%&			22.75\%&			21.60\%&			21.63\%&			21.68\%&			21.63\%
			  \\
			 \hline
			 \multirow{3}{*}{\textbf{ GWNet}} & MAE&\textbf{1.30}&			\underline{1.65}&			1.99&			1.59&			\underline{13.38}&			\underline{14.45}&			\underline{16.00}&			\underline{14.42}&			18.90&			20.48&			23.01&			20.47&			13.50&			\underline{14.41}&			\underline{15.81}&			\underline{14.40}
			 
			 \\ 
			 &RMSE&	
			 \underline{2.73}&			3.73&			4.60&			3.68&			\underline{23.25}&			\textbf{25.31}&			\textbf{27.73}&			\textbf{25.19}&			30.84&			33.50&			37.27&			33.47&			21.63&			\underline{23.45}&			\underline{25.77}&			\underline{23.39}
			  \\
			 &MAPE&2.71\%&			3.73\%&			4.71\%&			3.59\%&			13.82\%&			14.84\%&			16.03\%&			14.64\%&			7.89\%&			8.58\%&			9.81\%&			8.61\%&			\underline{8.63\%}&			\underline{9.24\%}&			\underline{10.11\%}&			\underline{9.21\%}
			  \\
			 \hline
			 \multirow{3}{*}{ \textbf{DCRNN}} &MAE&\underline{1.31}&			\underline{1.65}&			1.97&			1.59&			14.29&			15.19&			17.51&			15.53&			19.46&			21.13&			24.1&			21.16&			14.12&			15.23&			16.95&			15.22		
			 \\ 
			 &RMSE&	2.76&			3.75&			4.60&			3.69&			24.75&			26.55&			30.34&			27.18&			31.22&			34.14&			38.46&			34.14&			22.12&			24.18&			26.95&			24.17	
			  \\
			 &MAPE&		2.73\%&			3.71\%&			4.68\%&			3.58\%&			14.88\%&			15.33\%&			17.31\%&			15.62\%&			8.27\%&			8.95\%&			10.37\%&			9.02\%&			9.48\%&			10.21\%&			11.37\%&			10.21\%
			  \\
			 \hline
			 \multirow{3}{*}{\textbf{AGCRN}} &MAE&1.35&			1.67&			\underline{1.94}&			1.61&			14.22&			15.34&			16.86&			15.29&			19.25&			20.58&			22.68&			20.57&			14.26&			15.24&			17.03&			15.32		
			 \\ 
			 &RMSE&	2.88&			3.82&			\underline{4.50}&			3.70&			25.01&			26.99&			29.52&			26.95&			31.62&			34.39&			38.16&			34.40&			22.57&			24.40&			26.91&			24.41	
			  \\
			 &MAPE&		2.91\%&			3.81\%&			4.55\%&			3.64\%&			13.88\%&			15.86\%&			15.95\%&			15.15\%&			8.18\%&			8.69\%&			9.67\%&			8.74\%&			9.39\%&			9.85\%&			11.37\%&			10.03\%
			  \\
			 \hline
			 \multirow{3}{*}{ \textbf{STGCN}} &MAE&1.36&			1.70&			2.02&			1.64&			14.61&			15.57&			17.44&			15.65&			20.28&			21.68&			24.26&			21.74&			14.98&			15.97&			17.87&			16.08		
			 \\ 
			 &RMSE&	2.88&			3.84&			4.63&			3.76&			25.49&			27.27&			30.01&			27.31&			32.55&			35.15&			39.36&			35.27&			23.52&			25.30&			28.03&			25.39	
			  \\
			 &MAPE&		2.86\%&			3.79\%&			4.72\%&			3.67\%&			14.33\%&			15.07\%&			17.18\%&			15.39\%&			8.63\%&			9.18\%&			10.29\%&			9.24\%&			9.85\%&			10.51\%&			11.71\%&			10.60\%
			  \\
			 \hline
			 \multirow{3}{*}{ \textbf{MTGNN}} &MAE&1.33&			1.66&			1.95&			1.60&			13.71&			14.87&			16.44&			14.85&			19.32&			20.88&			23.45&			20.89&			14.16&			15.15&			16.78&			15.18		
			 \\ 
			 &RMSE&	2.80&			3.77&			4.50&			3.66&			\textbf{23.04}&			\underline{25.49}&			28.16&			\underline{25.32}&			31.23&			34.04&			38.08&			34.06&			22.37&			24.26&			26.83&			24.24	
			  \\
			 &MAPE&		2.81\%&			3.75\%&			4.62\%&			3.59\%&			14.36\%&			15.12\%&			16.01\%&		14.97\%&			8.31\%&			8.95\%&			10.18\%&			9.00\%&			9.67\%&			10.81\%&			11.53\%&			10.20\%
			  \\
			 \hline
			 \multirow{3}{*}{\textbf{ DGCRN}} &MAE&\textbf{1.30}&			\textbf{1.61}&			1.95&			\underline{1.57}&			13.46&			14.67&			16.36&			14.59&			\underline{18.70}&			\underline{20.22}&			\underline{22.46}&			\underline{20.17}&			\underline{13.44}&			14.52&			16.21&			14.42		
			 \\ 
			 &RMSE&	\textbf{2.72}&			\underline{3.67}&			4.54&			\underline{3.62}&			23.74&			26.06&			28.83&			25.92&			\textbf{30.59}&			\textbf{33.36}&			\underline{36.80}&			\underline{33.32}&			\underline{21.57}&			23.58&			26.37&			23.58	
			  \\
			 &MAPE&		\textbf{2.69\%}&			\textbf{3.59\%}&			\underline{4.54\%}&			\underline{3.48\%}&			14.53\%&			15.44\%&			16.84\%&			15.32\%&			\underline{7.81\%}&			\underline{8.39\%}&			\underline{9.44\%}&			\underline{8.42\%}&			\underline{8.63\%}&			9.40\%&			10.70\%&			9.46\%
			  \\
			 \hline
			 \multirow{3}{*}{\scriptsize \textbf{DFDGCN}} &MAE&\textbf{1.30}&			\textbf{1.61}&			\textbf{1.90}&			\textbf{1.55}&			\textbf{13.27}&			\textbf{14.33}&			\textbf{15.80}&			\textbf{14.28}&			\textbf{18.64}&			\textbf{20.11}&			\textbf{22.17}&			\textbf{20.03}&			\textbf{13.39}&\textbf{14.23}&\textbf{15.56}&\textbf{14.23}
			 \\ 
			 &RMSE&	\textbf{2.72}&			\textbf{3.66}&			\textbf{4.39}&			\textbf{3.57}&			23.47&			25.52&			\underline{27.92}&			25.39&			\underline{30.72}&			\underline{33.41}&			\textbf{36.70}&			\textbf{33.24}&			\textbf{21.53}&\textbf{23.34}&\textbf{25.65}&\textbf{23.30}
			  \\
			 &MAPE& \underline{2.70\%}&\underline{3.62\%}& \textbf{4.44\%} & \textbf{3.46\%} &\textbf{13.41\%}&\textbf{14.90\%}&\textbf{15.94\%}&\textbf{14.45\%}&\textbf{7.74\%}&\textbf{8.31\%}&\textbf{9.24\%}&\textbf{8.31\%}&\textbf{8.52\%}&\textbf{9.04\%}&\textbf{9.95\%}&\textbf{9.07\%} \\
			 \hline
		\end{tabular*}
		\vspace{-6mm}
\end{table*}
\begin{table}[t]
	\vspace{-3mm}
	\footnotesize
	\caption{Ablation experiments on graph convolution on PEMS08}
	\label{ablation}
	\centering
	\begin{tabular}{cc|cccc}
		\hline
		 \textbf{Graph} &  Metric& @3&@6&@12&Avg.\\
		 \hline
		 \hline
		  \multirow{3}{*}{\textbf{P}}&MAE&14.12&15.20&17.14&15.25\\
		 &RMSE&22.18&24.17&27.07&24.19\\
		 & MAPE&9.41\%&9.95\%&11.12\%&10.09\%\\
		 \hline
		  \multirow{3}{*}{\textbf{SA}}&MAE&13.54&14.47&15.88&14.45\\
		 &RMSE&\underline{21.63}&23.44&25.72&23.36\\
		 & MAPE&9.02\%&9.55\%&10.44\%&9.53\%\\
		 \hline
		 \multirow{3}{*}{\textbf{D}}&MAE&13.50&14.32&15.73&14.33\\
		 &RMSE&21.84&23.71&26.04&23.61\\
		 & MAPE&8.84\%&9.51\%&10.42\%&9.48\%\\
		 \hline
		 \multirow{3}{*}{\textbf{T}}&MAE&13.65&14.58&16.07&14.56\\
		 &RMSE&22.03&24.07&26.67&24.01\\
		 & MAPE&8.64\%&9.31\%&10.45\%&9.35\%\\
		 \hline
		 \multirow{3}{*}{\textbf{P+SA}}&MAE&13.50&			14.41&			15.81&			14.40\\
		 &RMSE&\underline{21.63}&			23.45&			25.77&			23.39\\
		 & MAPE&8.63\%&			9.24\%&			10.11\%&			9.21\%\\
		 \hline
		 \multirow{3}{*}{\textbf{D+P}}&MAE&13.51&14.34&15.80&14.36\\
		 &RMSE&21.68&23.54&26.05&23.52\\
		 & MAPE&8.87\%&9.42\%&10.43\%&9.46\%\\
		 \hline
		 \multirow{3}{*}{\textbf{D+SA}}&MAE&\textbf{13.32}&\textbf{14.17}&\textbf{15.45}&\textbf{14.16}\\
		 &RMSE&\textbf{21.53}&\textbf{23.24}&\textbf{25.33}&\textbf{23.15}\\
		 & MAPE&\underline{8.55\%}&\underline{9.10\%}&\underline{10.07\%}&\underline{9.13\%}\\
		 \hline
		 \multirow{3}{*}{\textbf{D+P+SA}}&MAE&\underline{13.39}&\underline{14.23}&\underline{15.56}&\underline{14.23}\\
		 &RMSE&\textbf{21.53}&\underline{23.34}&\underline{25.65}&\underline{23.30}\\
		 & MAPE&\textbf{8.52\%}&\textbf{9.04\%}&\textbf{9.95\%}&\textbf{9.07\%}\\
		 \hline
	\end{tabular}
\vspace{-8mm}
\end{table}

From Tab.\ref{res}, DFDGCN basically outperforms Baselines in the three metrics. Compared with GWNet, which is the base model of DFDGCN, the better results of DFDGCN indicate that the frequency domain graph actually captures the dynamic spatial dependence accurately. Compared with DGCRN which also mines dynamic spatial dependencies from traffic data, the rationality of our proposed problem and the validity of our model can be preliminarily judged.

Tab. \ref{ablation} presents the ablation experiment on graph convolution in our model to explore which graph plays a more important role in traffic prediction since both static predefined graphs as well as adaptive graphs are applied in DFDGCN. $P$ refers to predefined graphs and $SA$ refer to the self-adaptive graph, they are static graphs which preserve static spatial relationships in traffic network. T is called time-domain graph, which directly applies traffic data in the time domain for the graph construction in Section \ref{dfdgcn} without Fourier transform, and $D$ refers to our proposed frequency domain graph.

From the results of the convolution of a single graph, our dynamic frequency domain graph works the best, indicating that we effectively mine the dynamic spatial dependence. The comparison with T practically illustrates the serious impact of time-shift on data-driven spatial dependence mining and our effectiveness in analyzing spatial dependence from the frequency domain. When multiple graphs are convolved, the frequency-domain graph always brings more improvement. However, the effect always deteriorates when convolving the frequency domain graph with predefined graphs.
We conjecture that it may be attributed to these factors: (1). The static priori information of predefined graphs  play a negative role in traffic prediction; (2). Multiple graphs convolution may suffer from problems such as different information densities of individual graphs, inconsistency of convergence speed with the direction of hyperparameter gradient descent similar to that in multitask learning, which is a separate topic of research.

\vspace{-3mm}
\section{Conclusion}
\vspace{-3mm}
In this paper, we focus on spatial dependence in traffic prediction and propose a novel model DFDGCN. We argue that data-driven spatial dependence modeling suffers from time-shift and data noise, thus we propose to analyze traffic data in the frequency domain and construct an effective graph structure to mine spatial correlation in transportation network. Experiments support our argument and demonstrate that our model outperforms other baselines.
In subsequent research, we will focus on exploring spatial dependence between sensor signals and optimizing multi-graph convolution.

\bibliographystyle{IEEEbib}
\bibliography{strings,refs}

\end{document}